\documentclass[accepted]{el-mekkaoui_652} 

\usepackage[american]{babel}
\usepackage{subcaption}
\usepackage{algorithm}

\usepackage[noend]{algpseudocode}
\usepackage{comment}
\usepackage{textcomp}
\usepackage[utf8]{inputenc} 
\usepackage[T1]{fontenc}    
\usepackage{url}            
\usepackage{booktabs}       
\usepackage{amsfonts}       
\usepackage{nicefrac}       
\usepackage{microtype}      
\usepackage{amsmath, amsthm, amssymb}
\newtheorem{prop}{Lemma}
\newtheorem{ass}{Assumption}
\newtheorem{theorem}{Theorem}
\newtheorem{Remark}{Remark}
\usepackage{float}

\newcommand{\x}{\mathbf{x}}
\usepackage{graphicx}
\usepackage{bm}
\usepackage{xcolor}
\usepackage{multirow}
\usepackage{wrapfig}
\usepackage{natbib} 
\usepackage{mathtools} 
\usepackage{booktabs} 
\usepackage{tikz} 
\usepackage{todonotes}
\usepackage[nameinlink]{cleveref}

\title{Federated Stochastic Gradient Langevin Dynamics}

\author[1]{Khaoula el Mekkaoui\thanks{Equal contribution.}\thanks{Corresponding author: khaoula.elmekkaoui@aalto.fi}}
\author[1]{Diego Mesquita$^{*}$}
\author[2]{Paul Blomstedt}
\author[1,3]{Samuel Kaski}
\affil[1]{Helsinki Institute for Information Technology,Department of Computer Science,Aalto University, Finland}
\affil[2]{F-Secure, Finland}
\affil[3]{Department of Computer Science, University of Manchester, UK}

\begin{document}
\maketitle

\begin{abstract}
 Stochastic gradient MCMC methods, such as stochastic gradient Langevin dynamics (SGLD), employ fast but noisy gradient estimates to enable large-scale posterior sampling. Although we can easily extend  SGLD to distributed settings, it suffers from two issues when applied to federated non-IID data. First, the variance of these estimates increases significantly. Second, delaying communication causes the Markov chains to diverge from the true posterior even for very simple models. To alleviate both these problems, we propose \emph{conducive gradients},  a simple mechanism that combines local likelihood approximations to correct gradient updates. Notably, conducive gradients are easy to compute, and since we only calculate the approximations once, they incur negligible overhead. We apply conducive gradients to distributed stochastic gradient Langevin dynamics (DSGLD) and call the resulting method {federated stochastic gradient Langevin dynamics} (FSGLD).  We demonstrate that our approach can handle delayed communication rounds, converging to the target posterior in cases where DSGLD fails. We also show that FSGLD outperforms  DSGLD for non-IID federated data with experiments on metric learning and neural networks.  
\end{abstract}

\section{Introduction}

Gradient-based Markov Chain Monte Carlo (MCMC) methods are the de facto standard to sample from Bayesian posteriors. However, computing gradients exactly can be prohibitive even for moderately large data sets. Following the success of stochastic gradients in large-scale optimization and inspired by \citep{Welling+Teh:2011}, many MCMC algorithms were adapted to leverage fast but noisy gradient evaluations computed on mini-batches of data \citep{Ma+others:2015}. Examples include stochastic gradient Langevin dynamics (SGLD) \citep{Welling+Teh:2011}, stochastic gradient Hamiltonian Monte Carlo (SGHMC) \citep{Chen+others:2014}, and Riemann manifold Hamiltonian Monte Carlo (RMHMC) \citep{RMHMC2001}. These methods have established themselves as popular choices for scalable Bayesian inference.

\begin{figure}[tp] 
    \centering
    \includegraphics[width=0.8\linewidth]{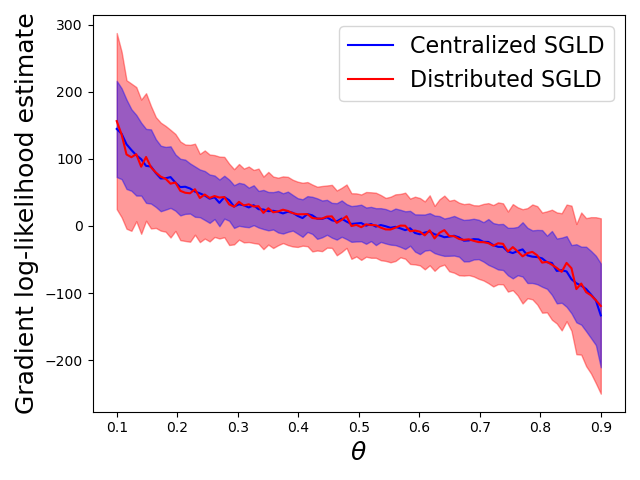}
    \caption{Comparison between gradient estimators using centralized (SGLD) and distributed data (DSGLD), for a model with Bernoulli likelihood and uniform prior. We computed gradients using 5 samples from a total of 30 observations generated from fair coin tosses. For DSGLD, we simulate the federated non-IID regime splitting data between 3 equally-available shards of same size but distinct means -- $0.1$, $0.5$ and $0.9$. The confidence bars reflect one standard deviation.
    DSGLD {(red)} shows higher variance than SGLD {(blue)} even for this simple case. 
    } 
    \label{fig:variance_motivation}
\end{figure}

Complementary to data subsampling, which underlies the use of stochastic gradients,  we can also split data across several workers and use distributed computations to scale up MCMC \citep{Neiswanger+others:2014, Scott+others:2016,  Terenin+others:2015, Wang+others:2015, Nemeth2018, Mesquita2019}. 
In particular,  \citet{Ahn+others:2014}  extend stochastic gradient MCMC using a simple estimator that accounts for data partitions and propose distributed SGLD (DSGLD). More specifically, DSGLD operates passing a Markov chain between computing nodes and using only local data to estimate gradients at each step. The gradient estimates are then scaled accordingly to correct the bias.

Despite the popularity of stochastic gradient MCMC, no work has considered applications to federated learning. Notably, federated data arise independently in different clients/devices, and communication or privacy constraints prevent it from being disclosed to a server. As a consequence, data are often partitioned in a non-IID fashion. We argue that distributed methods such as vanilla DSGLD are inappropriate for the non-IID regime. In practice, this regime significantly amplifies the variance of stochastic gradients. \autoref{fig:variance_motivation} illustrates this phenomenon for a simple model with Bernoulli likelihood. In turn, this can lead to poor mixing rates and slow convergence\citep{Dubey+others:2016}.  Additionally, in federated settings, we want to avoid frequent communication, which can introduce bias and cause DSGLD to diverge from the target posterior \citep{Ahn+others:2014}.

To mitigate these problems, we propose \emph{conducive gradients},  a zero-mean stochastic function that aggregates approximations of each client's likelihood factor. These approximations are computed independently on the client-side before and communicated only once.
We add our conducive gradient to the gradient estimator proposed by \citet{Ahn+others:2014} to derive a novel method, which we call federated stochastic gradient Langevin dynamics (FSGLD). We demonstrate that i) FSGLD converges to the true posterior in cases where DSGLD fails, and ii) FSGLD outperforms DSGLD in federated non-IID scenarios.

In general, we can compute conducive gradients in constant time and, since we compute the local approximations only once, FSGLD has the same computational complexity as DSGLD.  We also provide convergence bounds for FSGLD and use these results to gain insight regarding how to efficiently choose local likelihood approximations which minimize the bound. Furthermore, we provide analysis for DSGLD since no formal analysis is available in the literature. There are well-established analyses of convergence for SGLD in serial settings\citep{Chen:2015, Nagapetyan2017, Baker2019, Teh+ohters:2016, Vollmer+others:2016}, which we use as a starting point due to their relatively straightforward formulation.

We organize the remainder of this work as follows.
In \Cref{sec:background}, we establish the notation and provide a brief review of serial and distributed SGLD.
In \Cref{sec:cg}, we introduce the concept of conducive gradients and use it to derive a novel SGLD algorithm tailored to federated data.
In \Cref{sec:analysis}, we show convergence bounds for both our method and DSGLD.
In  \Cref{sec:examples}, we show experimental results. 
Finally, we discuss related work in \Cref{sec:related} and draw conclusions in Section~\ref{sec:discussion}.

\section{Background and notation}\label{sec:background}

Let $\x = \{x_1,\ldots,x_N\}$ be a data set of size $N$ and let $p(\theta|\x)\propto p(\theta)\prod_{i=1}^N p(x_i|\theta)$ be the density of a posterior distribution from which we wish to draw samples. Langevin dynamics \citep{Neal:2011} is a family of MCMC methods which utilizes the gradient of the log-posterior, 
\[
\nabla \log p(\theta|\x) = \nabla \log p(\theta) + \sum_{i=1}^N \nabla \log p(x_i|\theta),
\]
to generate proposals in a Metropolis-Hastings sampling scheme. For large data sets, computing the gradient of the log-likelihood with respect to the entire data set $\mathbf{x}$ becomes expensive. To mitigate this problem, stochastic gradient Langevin dynamics (SGLD) \citep{Welling+Teh:2011} uses stochastic gradients to approximate the full-data gradient.

Denoting by $\nabla \log p ( \x^{(m)} | \theta_t ) = \sum_{x\in \x^{(m)}} \nabla \log p ( x | \theta_t )$ the gradient of the log-likelihood with respect to a minibatch $\x^{(m)}$ of size $m$, SGLD draws samples from the target distribution using a stochastic gradient update of the form
\begin{equation}\label{eq:sgld_update}
 \theta_{t+1} = \theta_{t} +
  \frac{h_t}{2} v(\theta_t) + \eta_t,
\end{equation}
in which $h_t$ is the step size, $\eta_t$ is a noise variable sampled from $\mathcal{N}(0,h_t I)$ and where
\begin{equation}
    v(\theta_t) =  \nabla \log p ( \theta_t )  + \frac{N}{m}\nabla \log p ( \x^{(m)} | \theta_t ).
\end{equation} 
In theory, step size is annealed according to a schedule satisfying $\sum_{t=1}^\infty h_t = \infty$ and  $\sum_{t=1}^\infty h_t^2 < \infty$. Note that as $h_t \rightarrow 0$, the Metropolis-Hastings acceptance rate goes asymptotically to one and thus, the accept-reject step is typically ignored in SGLD. While a proper annealing schedule yields an asymptotically exact algorithm, constant step sizes are commonly used in practice.

\subsection{DSGLD}

\citet{Ahn+others:2014} propose DSGLD as an extension of SGLD to distributed settings (DSGLD),
where we find data $\mathbf{x}$ to be partitioned into $S$ non-overlapping shards $\x_s$, each held by a client, such that $\x = \{\x_1,\ldots,\x_S\}$.
More specifically, DSGLD uses a modified version of the update in \Cref{eq:sgld_update} which is better suited for distributed data.
The main idea is that in each iteration, a mini-batch is sampled \emph{within} a shard, say $\mathbf{x}_s$, and the shard itself is sampled by a scheduler with probability $f_s$, with $\sum_{s=1}^S f_s =1$ and $f_s > 0$ for all $s$. This results in the update 
\begin{equation}\label{eq:dsgld}
\theta_{t+1} = \theta_t + \frac{h_t}{2}v_{s_t}(\theta_t) 
 + \eta_t,  
\end{equation}
in which $v$ is an unbiased gradient estimator given by
\begin{equation}
    v_{s_t}(\theta_t) = \nabla \log p(\theta_t) + \frac{N_{s_t}}{f_{s_t}m} \nabla \log p( \x_{s_t}^{(m)} | \theta_t),
\end{equation}
and where $N_{s_t}$ denotes the size of shard $\x_{s_t}$, chosen at time $t$.  
Intuitively, if a mini-batch of $m$ data points is chosen uniformly at random from $\x_{s_t}$, then $N_{s_t}/m$ scales $\nabla \log p(\x_{s_t}^{(m)} | \theta_t)$ to be an unbiased estimator for $\nabla \log p(\x_{s_t} | \theta_t)$ , while $f_{s_t}^{-1}$ further scales this gradient to be an unbiased estimator for  $\nabla \log p(\x | \theta_t)$.

A downside of this approach is the constant communication between workers. To alleviate this problem,  \cite{Ahn+others:2014} propose taking multiple update steps within the same shard before moving to another worker. Nonetheless, this reduction in communication costs comes at the expense of some loss in asymptotic accuracy.  It is worth noting that, while the data are distributed, we can still understand  DSGLD chains as entirely serial procedures. In practice, however, distributed settings are naturally amenable to running multiple chains in parallel.

\section{FSGLD: Federated inference using conducive gradients}\label{sec:cg}
In DSGLD, stochastic gradient updates are computed on mini-batches sampled within the data shard of a specific client, which adds bias to the updates and increases variance globally. This is especially significant if for non-IID federated data, when shards are heterogeneous and likelihood contributions vastly differ between two devices. 
To counteract this, we would like to make the local updates benefit from other shards' information, ideally without significantly increasing either computational cost or memory requirements.
Our strategy to achieve this goal is to augment the local updates, in \Cref{eq:dsgld}, with an auxiliary gradient computed on a tractable surrogate  for the full-data likelihood $p(\x|\theta)$.

We assume here that the surrogate, denoted as $q(\theta) $, factorizes over shards, such as $q(\theta) = \prod_s q_{s}(\theta)$, where each $q_{s}(\theta)$ is itself a surrogate for $p(\x_{s} | \theta)$, i.e. the likelihood w.r.t. an entire shard $s$. 
Given these surrogates, we define the \textit{conducive gradient} w.r.t. shard $s$ as 
\begin{align*}
     g_s(\theta) &= \nabla \log q(\theta) - \frac{1}{f_{s}} \nabla \log q_{s}(\theta).
\end{align*}
Using the conducive gradient $g_s$ we define our novel update rule as
\begin{equation}
\begin{split}
    \theta_{t + 1} =
    \theta_t &+  \frac{h_t}{2} \bigg( v_{s_t}(\theta_t)   
      +  g_{s_t}(\theta_t) \bigg)  + \eta_t.
\end{split}
   \label{eq:cg_update}
\end{equation}

Algorithm \ref{alg:CGDGLD} describes our method, federated stochastic Langevin dynamics (FSGLD). We discuss the validity of our novel gradient estimator ($v_{s} + g_{s}$) in \Cref{sec:analysis}.

\begin{Remark}[\textbf{Controlling exploration}]
Note that conducive gradients can alternatively be written as 
\begin{equation}
    g_s(\theta) = \nabla \log \frac{q (\theta)}{ q_s(\theta)^{f_s^{-1}} },
\end{equation}
making it explicit that these terms encourage the exploration of regions in which we believe, based on the approximations $q$ and $q_s$, the posterior density to be high but the density within shard $s$ to be low. 
We can explicitly control the extent of this exploration by multiplying the conducive gradient by a constant $\alpha>0$ to obtain the modified gradient estimator:
\begin{equation}
        \frac{N_{s}}{f_{s}m}\nabla \log p ( \x_s^{(m)} | \theta ) + \alpha g_s(\theta).
\end{equation}
\end{Remark}

\begin{algorithm}[t]
\caption{FSGLD}
\label{alg:CGDGLD}
\begin{algorithmic}
    \State \textbf{Client-side} \textit{Update}($T$, $\theta_0$, $s$)
    \State \textbf{\textit{Given:}} Total number of iterations $T$, step sizes $\{h_t\}_{t=0}^{T-1}$, initial chain state $\theta_0$, client number $s$.
    \For{ $t = 0\ldots T-1$}
        \vspace{0.5em}
        \State Sample a mini-batch $\x_s^{(m)}$ of size $m$ from $\x_s$
        \State $d_s \leftarrow \nabla \log p(\theta) + \frac{1}{f_{s}} \frac{N_{s}}{m} \nabla \log p ( \x_{s}^{(m)} | \theta ) $ 
        
        \vspace{0.5em}
        \Comment{ DSGLD estimator}
        \vspace{1mm}
        \State $ g_s  \leftarrow - \frac{1}{f_{s}} \nabla \log q_{s}(\theta) + \nabla \log q(\theta)$
        
        \vspace{0.5em}
        \Comment{Conducive gradient}
        \State $\theta_{t+1} \leftarrow \theta_t + \frac{h_t}{2} (d_s + g_s) + \eta_t$
        
        \vspace{0.5em}
        \Comment{ According to Eq.~(\ref{eq:cg_update})} 
        \vspace{1mm}
    \EndFor \textbf{end for}
    \State Reassign\_chain($\theta_0, \ldots, \theta_T $)
    
        \vspace{0.5em}
    \Comment{Send chain to server}

    \vspace{1em} 
    \State \textbf{Server-side} Reassign\_chain($\theta_0,\ldots,\theta_{T}$)  
    \State \textbf{\textit{Given:}} Client probabilities $f_1, \ldots, f_S$. 

    \State Store the received chain 
    \State $c \sim \texttt{Categorical}(f_1, \ldots, f_S)$
     
     \vspace{0.5em}
    \Comment{Choose a client at random.}
    \State \textit{Update}($T$,  $\theta_{T}$, $c$)
         
    \vspace{0.5em}
    \Comment{Pass on chain to client $c$.}
\end{algorithmic}
\end{algorithm}

\subsection{Choice of surrogates $q_1, \ldots, q_S$}
\label{subsec:choice}


The key idea in choosing $q_s$ is to obtain an approximation of $p(\x_s|\theta)$ with a parametric form,
and to choose the parametric form such that $\nabla \log q_s(\theta)$ computation is inexpensive.
It is particularly beneficial if the gradient can be computed in a single gradient evaluation instead of iterating over all data, of size $N_s$, of said shard $s$.
Exponential family distributions  are especially convenient for this purpose, as they are closed under product operations, enabling us to compute $\nabla \log q(\theta_t)$ in a single gradient evaluation. This keeps the additional cost of our method negligible even when $S \gg m$. 
More specifically, with exponential family surrogates, this cost is  constant with respect to the number of clients $S$. 

In this work, we use a simulation-based approach to compute $q_s$ by first drawing from $p_s \propto p(\x_s|\theta)$ locally employing SGLD, and using the resulting samples to compute the parameters of an exponential family approximation.
To avoid communication overhead, $q_1, \ldots, q_S$ can be computed independently in parallel for each of the data shards and then communicated to the coordinating server once, before the FSGLD steps take place.

\section{Analysis}\label{sec:analysis}
In this section, we analyze the convergence of both DSGLD  and the proposed FSGLD. While simple, our results reflect the impact of heterogeneity among data shards and, additionally, our bounds for FSGLD also provide deeper insight on our strategy for choosing the surrogates $q_1, \ldots, q_S$.
We provide proofs in the supplementary material.

\subsection{Convergence of DSGLD}\label{sec:convergence}

We begin by analyzing the convergence of DSGLD under the same framework used for the analysis of SGLD by \citep{Chen:2015} and subsequently adopted by \citep{Dubey+others:2016}, who directly tie convergence bounds to the variance of the gradient estimators.
Besides certain regularity conditions (see Appendix A) adopted in these works, which we outline in the supplementary material, we make the following assumption:

\begin{ass} The gradient of the log-likelihood of individual elements within each shard is bounded, i.e., 
    $ \| \nabla \log p(x_i | \theta )\| \leq \gamma_s$, for all $\theta$ and $x_i \in \x_s$, and each  $s \in \{1, \ldots, S\}$.

    \label{ass:1}
\end{ass}

We then proceed to derive the following bound on the convergence in mean squared error (MSE) of the Monte Carlo expectation $\hat{\phi} = T^{-1} \sum_{t=1}^T \phi(\theta_t) $ of a test function $\phi$ with respect to its expected value $\overline{\phi} = \int \phi(\theta) p(\theta|\x) \, d \theta$.

\begin{theorem}
Let $h_t = h\,$ for all  $t \in \{1, \ldots, T\}$. Under standard regularity conditions and Assumption \ref{ass:1}, the MSE of DSGLD for a smooth test function $\phi$ at time $K = h T$ is bounded, for some constant $C$ independent of  $T$ and $h$, in the following manner:
    \[
        \mathbb{E}\left(\hat{\phi} - \overline{\phi}\right)^2 \leq C \Bigg( \frac{  \sum_s \frac{N_s^2}{f_s}  \gamma_s^2}{m T} + \frac{1}{h T} + h^2 \Bigg) .
    \] \label{bound1}
\end{theorem}

The bound in Theorem \ref{bound1} depends explicitly on the ratio between squared shard sizes and their selection probabilities. This follows the intuition that both shard sizes and their availability play a role in the convergence speed of DSGLD. 

\begin{Remark}[\textbf{SGLD as a special case}]
\label{remark:special_case}Note also that bound for DSGLD generalizes previous results for SGLD \citep{Dubey+others:2016}. More specifically, if we combine all shards into a single data set $\x = \cup_s \x_s$ and let $\gamma \geq \gamma_1, \ldots, \gamma_s$, we recover the bound for SGLD:
\[
\mathbb{E}\left(\hat{\phi} - \overline{\phi}\right)^2 
\leq C \left( \frac{ {N^2}  \gamma^2  }{m T} + \frac{1}{h T} + h^2 \right).
\]
\end{Remark}

Note that the constant $\gamma$ in the remark above is an upper-bound on the gradient log-likelihood for any specific data point in $\x$. 

\subsection{Convergence of FSGLD}
The following result states that when $g_s(\theta)$ is added to the stochastic gradient in a DSGLD setting, the resulting estimator remains a valid estimator for the gradient of the full-data log-likelihood.
\begin{prop}
    Assume $\log q_1, \ldots, \log q_S$ are Lipschitz continuous.
    Given a data set $\x$ partitioned into shards $\x_1, \ldots, \x_S$, with respective sample sizes $N_1, \ldots, N_S$ and shard selection probabilities $f_1, \ldots, f_S$,
    the following gradient estimator,
        \begin{equation*}
        \frac{N_{s}}{f_{s}m}\nabla \log p ( \x_s^{(m)} | \theta ) + g_s(\theta),
        \end{equation*}
    is an unbiased estimator of $\nabla \log p ( \x| \theta )$ with finite variance.
    \label{prop:lemma1}
\end{prop}

With the validity of our estimator established, we now provide the convergence bound for FSLGD, stated in the following theorem.
\begin{prop}
   If $\,\log q_1, \ldots, \log q_S$ are everywhere differentiable and Lipschitz continuous, then the average value of $\|\nabla \log p(x_i | \theta) - N_s^{-1} \nabla \log q_s(\theta)\|^2$, taken over $x_i \in \x_s$, is bounded by some  $\epsilon_s^2$, for each $\theta$. \label{prop:3} 
\end{prop}

\begin{theorem}
    Let $h_t = h$ for all  $t \in \{1, \ldots, T\}$. Assume $\log q_1, \ldots, \log q_S$ are Lipschitz continuous. Under standard regularity conditions (Appendix A) and Assumption \ref{ass:1}, the MSE of FSGLD(defined in Algorithm \ref{alg:CGDGLD}) for a smooth test function $\phi$ at time $K = h T$ is bounded, for some constant $C$ independent of $T$ and $h$ in the following manner:
    \small
    \begin{equation*}
    \begin{split}
        \mathbb{E}&\left(\hat{\phi} - \overline{\phi}\right)^2 \leq 
        C \Bigg( \frac{1}{m T} { \sum_s \frac{N_s^2}{f_s}  \epsilon_s^2}  
        + \frac{1}{h T} + h^2 \Bigg) .
    \end{split}
    \end{equation*} \label{prop:conv_cgdsgld}
    \normalsize
\end{theorem}

In other words, Theorem \ref{prop:conv_cgdsgld} tells us that we can counteract the effect of data heterogeneity by choosing surrogates $q_s$'s that make the constants $\epsilon_s^2$'s as small as possible.

\begin{Remark}[\textbf{Revisiting the choice of $q_s$'s}] Naturally, the choice of $q_s$ exherts direct influence on $\epsilon_s^2$. Doing so analytically is difficult, but we can get further insight if we choose $q_s$ that achieves
\begin{equation}
\min_{q_s} \max_\theta \| \nabla \log p(\x_s | \theta) - \nabla \log q_s(\theta) \|^2,
\label{eq:max_upper_bound}
\end{equation}
which itself minimizes an upper-bound for $\epsilon_s^2$.
Note that \autoref{eq:max_upper_bound} reaches its minimum when $q_s(\theta)$ is equal to $p(\x_s|\theta)$.
Therefore, it is sensible to choose $q_s$ that approximates the local likelihood contributions as well as possible, as previously described in Subsection \ref{subsec:choice}.
We provide further details about the upper-bound in the supplementary material.
\end{Remark}


\section{Experiments}\label{sec:examples}

\begin{figure*}[ht!]
    \centering
    \includegraphics[width=0.9\textwidth]{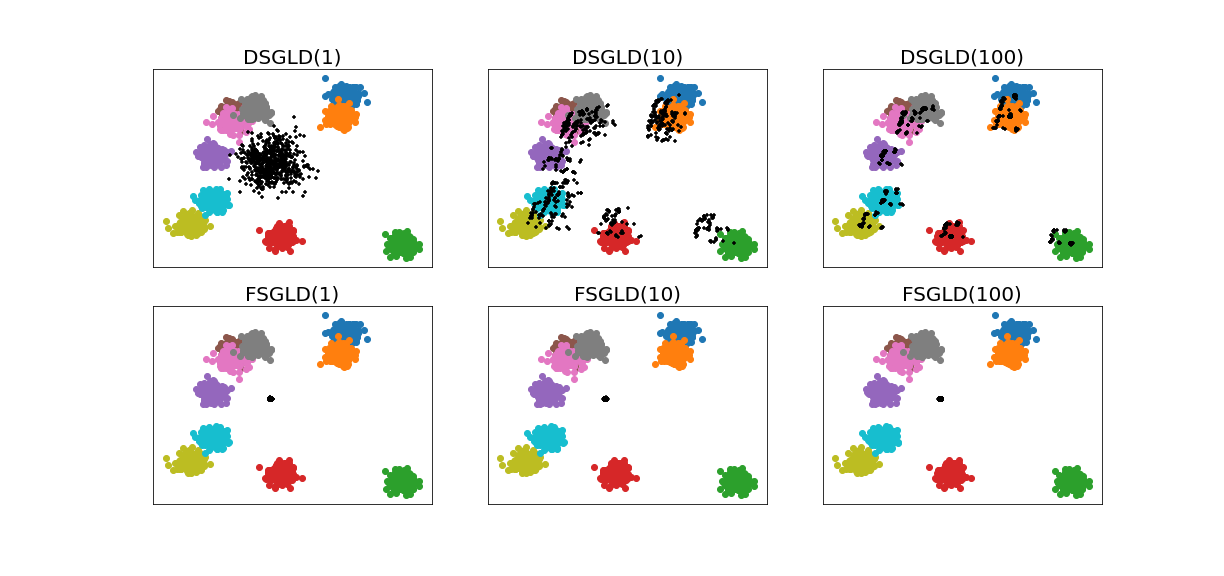}
    \caption{Posterior samples of the global mean (black) in DSGLD and FSGLD, as a function of the number of shard-local updates (shown in parentheses in each title). The colored dots are the data samples, different shards having different color. FSGLD converges to the target posterior in all cases, while DSGLD approaches a mixture of local posteriors as the number of local updates increases. For reference, \autoref{fig:blob_posterior} shows samples from the analytical posterior.}
    \label{fig:blobs}
\end{figure*}

In this Section, we demonstrate the performance of our method (FSGLD) for increasingly complex models under the  non-IID data regime, which is a defining characteristic of federated settings. 

In Subsection \ref{exp:hetero_shards}, we show that DSGLD quickly diverges from the true posterior as the frequency of communication decreases, even for very simple models.
On the other hand FSGLD easily circumvents this pathology with the help of conducive gradients. 
In Subsection \ref{exp:metric}, we consider the task of inferring Bayesian metric learning posteriors from highly non-IID data. 
Finally, in Subsection \ref{exp:bnn}, we  show how our method can be employed to learn Bayesian neural networks in a distributed fashion, comparing it against DSGLD for both federated IID and non-IID data.

While these models are progressively more complex, we highlight that, using simple multivariate Gaussians as the approximations $q_1, \ldots, q_S$ to each client's likelihood function, we are still able to obtain good and computationally scalable results for FSGLD. For the first set of experiments, we derive analytic forms for the approximations $q_s$, which is only possible due to the simplicity of the target model. For the remaining experiments, we employ SGLD independently for $s=1,\ldots,S$ and use the samples obtained to compute the mean vector and covariance matrix that parameterize $q_s$. We implemented all experiments using PyTorch~\footnote{\href{https://pytorch.org}{https://pytorch.org}}. 
We also show additional experiments in the supplementary material. %

It is important to highlight that, due to our choice of exponential family surrogates, the cost of evaluating conducive gradients in the following experiments compares to an additional prior evaluation.

\subsection{non-IID data and delayed communication} \label{exp:hetero_shards}

An ideal sampling scheme would, in theory, update the chain once at a device and immediately pass it over to another device in the next iteration. However, 
such a short communication cycle would result in a large overhead and is unrealistic in federated settings. 
To ameliorate the problem, \cite{Ahn+others:2014} proposed making a number of chain updates before moving to another device.
However, the authors reported that as the number of iterations within each client and shard increases, 
the algorithm tends to lose sample efficiency and effectively sample from a mixture of local posteriors, 
$\frac{1}{S}\sum_{s=1}^S p(\theta|\x_s)$,
instead of the true posterior. 
With heterogeneous data shards, the effect is particularly noticeable.
In this experiment, we illustrate the pathology and  show how FSGLD overcomes it.

\textbf{Model:}
We consider inference for the mean vector $\mu$ of normally distributed data under the simple model
\begin{equation}
    p(\mu | \x) \propto \mathcal{N}(\mu | \mathbf{0}, I) \prod_{s=1}^{S} \mathcal{N}(\x_s | \mu, I).
    \label{eq:simpmdl}
\end{equation}

\textbf{Setting:}
We generate $S=10$ disjoint subsets $\x_1, \ldots, \x_S$ of size $200$, each respectively from $\mathcal{N}(\mu_1, I)$ with $\mu_1$ uniformly sampled from the $[-6, 6] \times [-6, 6]$ square (the colored dots in Fig \ref{fig:blobs}). We then perform inference on the overall mean $\mu$ using the model (\ref{eq:simpmdl}). We sample the same number of posterior samples using both DSGLD and FSGLD with fixed step-size $h = 10^{-4}$, mini-batch size $m=10$ and $f_1 = \cdots = f_S = 1/S$. The first $20000$ samples were discarded, and the remaining ones were thinned by 100. 

\textbf{Choice of $q_s$'s:}
We use analytic Gaussian surrogates $q_s(\theta) = \mathcal{N}(\theta |\overline{\x}_s , N^{-1}I)$, for each $s=1\ldots S$. Note that $q_s$ here is exactly likelihood function induced by the shard $\x_s$. 

\begin{figure}[htp]
\centering
    \begin{subfigure}[]{0.4\textwidth}
        \centering
        \includegraphics[width=\linewidth]{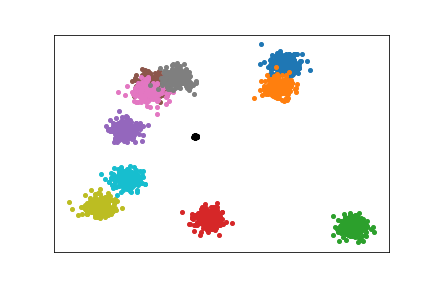}
        \caption{}\label{fig:blob_posterior}
    \end{subfigure}
    \begin{subfigure}[]{0.4\textwidth}
        \centering
        \includegraphics[width=\linewidth]{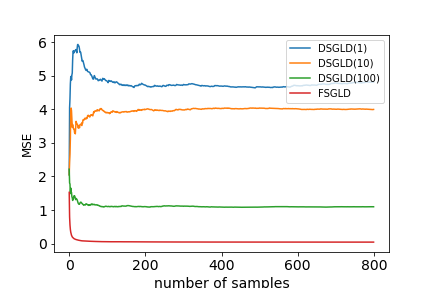}
        \caption{}\label{fig:mse_blobs}
    \end{subfigure}
    \caption{(a) Samples from the analytical posterior, for comparison with Figure 2. (b) Quantitative comparison of MSE $|\bar{\mu} - \hat{\mu}|^2$ in estimating the global mean, as a function of the number of posterior samples, showing FSGLD clearly outperforms DSGLD. Numbers in parentheses indicate number of successive local shard updates; for FSGLD the curves are almost identical, independently of the number, and only one is shown. Only FSGLD converges to  $\bar{\mu}$.}
    \label{fig:blob_sorted}
\end{figure}

\textbf{Results:}
Figure \ref{fig:blobs} shows the posterior samples as a function of the number of local updates the method takes before passing the chain to the next device. For comparison, Figure \ref{fig:blob_posterior} shows samples from the analytical posterior. As we can see from the results, the proposed method (FSGLD) converges adequately to the true posterior while DSGLD diverges towards a mixture of local approximations. 
This discrepancy becomes more prominent as we further increase the number of local updates, therefore delaying communication.
Figure \ref{fig:mse_blobs} compares the convergence in terms of the number of posterior samples. While DSGLD with a hundred local updates converges as fast as FSGLD, it plateaus at a higher MSE. Note that in contrast with DSGLD, FSGLD is insensitive to the number of local updates in the current experiment.

\begin{figure}[htp]
    \centering
    \includegraphics[width=0.45\textwidth]{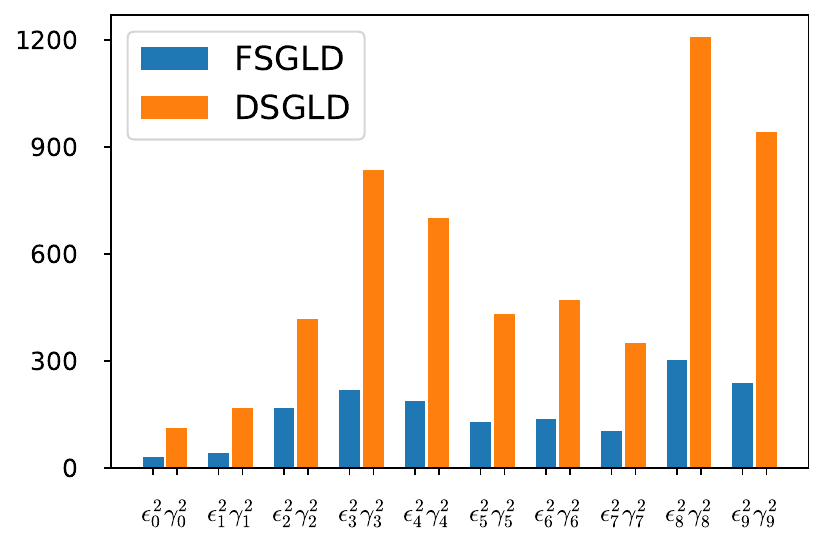}
    \caption{The constants $\epsilon_s^2$' s, which govern our method (FSGLD), are much smaller than $\gamma_s^2$' ’s from DSGLD, resulting in tighter bounds. The bar plot shows grid-based approximations these for these values in the mean estimation model, in \Cref{eq:simpmdl}.}
    \label{fig:epsilonvsgamma}
\end{figure}

\paragraph{Quantifying $\epsilon_s^2$'s.} \autoref{prop:conv_cgdsgld} states that the upper-bound on the MSE of FSGLD deteriorates as $\epsilon_1^2, \ldots, \epsilon_S^2$ increase. While computing $\epsilon_s^2$'s is intractable for most models, we can approximate it in this simple case using a grid. The same can be done for  $\gamma_s^2$'s, which govern the DSGLD bound in a similar manner. Figure \ref{fig:epsilonvsgamma} shows that $\epsilon_s^2 \ll \gamma_s^2 $ for all shards $s$. This phenomenon also corroborates with the results in \autoref{fig:mse_blobs}, which show that the MSE of FSLGD converges to much smaller values than DSGLD.

\subsection{Metric learning}\label{exp:metric}


Given sets of similar $\mathcal{S}$ and dissimilar $\mathcal{D}$ pairs of vectors from $\mathcal{X} = \{ x_n \}_{n=1}^N \in \mathbb{R}^D$, metric learning concerns the task of learning a distance metric matrix $A \in  \mathbb{R}^{D\times D}$ such that the Mahalanobis distance
\[
    \|x_i - x_j \|_A = \sqrt{(x_i - x_j)^\intercal A (x_i - x_j)}
\]
is low if $(x_i, x_j) \in \mathcal{S}$ and high if $(x_i, x_j) \in \mathcal{D}$.

\textbf{Model:}
We consider inference on the Bayesian metric learning model proposed by \cite{Yang2012}, in which it is assumed that $A$ can be expressed as $\sum_k \gamma_k \mathbf{v}_k \mathbf{v}_k^\intercal$ where $\mathbf{v}_1, \ldots, \mathbf{v}_K$ are the top $K$ eigenvectors of $X = [x_1^\intercal, \ldots, x_N^\intercal]$. 
The likelihood function for each pair $(x_i, x_j)$ from $\mathcal{S}$ or $\mathcal{D}$ is given by 
\begin{align*}
\begin{split}
    y_{ij} | (x_i, x_j), A, \mu \sim&  \text{Ber} \left( \frac{1}{1 + \exp\left( y_{ij} \left( \| x_i - x_j \|_A^2 - \mu \right) \right) }  \right) 
\end{split}
\end{align*}

where $y_{ij}$ equals one if $(x_i, x_j) \in \mathcal{S}$ and equals zero, if $(x_i, x_j) \in \mathcal{D}$.  
While having $\bm{\gamma} = [\gamma_1, \ldots, \gamma_K] \succ 0$ is enough to guarantee that $A$ is positive definite and defines distance metric, this requirement is relaxed and a diagonal Gaussian prior is put both on $\bm{\gamma}$ and $\mu$. For a more thorough treatment, we refer the reader to the original work by \citet{Yang+others:2016}.

\textbf{Setting:} 
We have devised a dataset for metric learning based on the Spoken Letter Recognition\footnote{\url{https://archive.ics.uci.edu/ml/datasets/isolet}} (isolet) data, which encompasses 7797 examples split among 26 classes.  We have created $|\mathcal{S}| = 5000$ and $|\mathcal{D}| = 5000$ pairs of similar and dissimilar vectors, respectively, using the labels of the isolet examples, i.e., samples are considered similar if they belong to the same class and dissimilar otherwise.  To obtain a federated non-IID dataset, we split these pairs into $S=10$ data shards of identical size, in a manner that there is no overlap in the classes used to create the sets of pairs $\mathcal{S}_s$ and $\mathcal{D}_s$ in each shard $s=1\ldots S$. Additionally, we created another thousand pairs of equally split similar and dissimilar examples that we hold out for the test.

\textbf{Choice of $q_s$'s:}
We set $f_1 = \ldots = f_S = \frac{1}{S}$ and run both DSGLD and FSGLD with constant step size $10^{-3}$ and mini-batch size $m=256$. We use simple Gaussian approximations for $q_1, \ldots, q_S$, with mean and covariance parameters computed numerically using three thousand samples from
\begin{align*}
    \prod_{(x_i, x_j) \in \mathcal{S}_s \cup \mathcal{D}_s} 
\text{Ber}   \left( y_i \big| \frac{1}{1 + \exp\left( y_{ij} \left( \| x_i - x_j \|_A^2 - \mu \right) \right) }  \right),
\end{align*}
obtained running SGLD independently for each client. 
 It is worth highlight that these approximations are only computed once at the beginning of training, and then communicated to the server. 
Appendix F.2 shows additional results using Laplace approximations.

\begin{figure}[t!]
\centering
    \begin{subfigure}[]{0.4\textwidth}
        \centering
        \includegraphics[width=\linewidth]{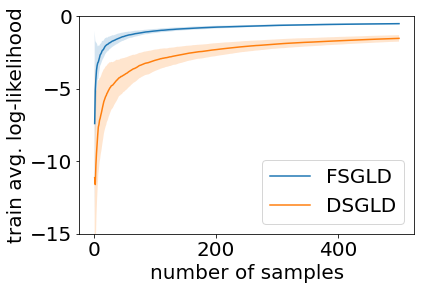}
        \caption{}
    \end{subfigure} \hfill
    \begin{subfigure}[]{0.4\textwidth}
        \centering
        \includegraphics[width=\linewidth]{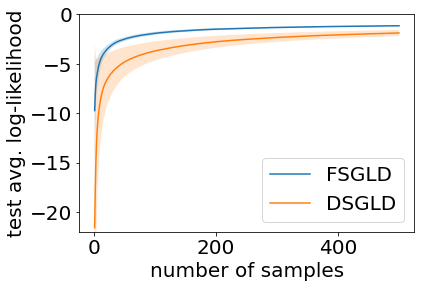}
        \caption{}
    \end{subfigure}
    \caption{\textbf{Metric learning.} Average log-likelihood values as a function of the number of samples for both DSGLD and FSGLD, measured on (a) learning data and (b) on a held-out set of test samples.The curves show the mean and standard deviation (error bars) for ten repetitions of the experiment, with different random seeds. FSGLD converges faster and to better predictions than DSGLD, both for training data and for test data. }
    \label{fig:metric}
\end{figure}

\textbf{Results:}  
Figure \ref{fig:metric} shows results in terms of average log-likelihood as a function of the number of samples. The results show that $i$) FSGLD achieves better performance than DSGLD both in learning and on test data; and $ii$) FSGLD is more communication-efficient than DSGLD, i.e., it converges faster. Furthermore, FSGLD exhibits overall lower standard deviation.

\subsection{Bayesian neural networks}\label{exp:bnn}
We now gauge the performance of the proposed FSGLD for posterior inference on a deep Multi-Layer Perceptron (MLP) both for federated IID and for federated non-IID data. 

\textbf{Model:} We consider an MLP with three hidden layers. The first two hidden layers consist of $18$ nodes each, and the last one of $8$. We equip all hidden nodes with the rectified linear unit (ReLU) activation function, and we apply the Softmax function to the output of the network. Since we employ this network for a classification task, we use the cross-entropy loss function.

\textbf{Setting:} For this experiment, we create a series of $S=30$ label-imbalanced data shards based on the SUSY\footnote{\url{https://archive.ics.uci.edu/ml/datasets/SUSY}} dataset.
For the IID case, we draw the proportions $\pi_1, \ldots \pi_S$ of positive samples in each shard $s=1,\ldots, S$ from a symmetric Beta distribution with parameters $a = b = 100$. In this case, each data shard is approximately balanced.
For the non-IID case, we sample the proportions from a Beta with both parameters equal to $0.5$ instead.
When $a=b=0.5$, half of the shards tend to have mostly positive and the other half tends to have mostly negative labels, enforcing diversity between  shards.

All the shards mentioned above comprise $9\times10^4$ samples, and we hold out an additional balanced shard for evaluation.
We ran FSGLD and DSGLD, with even shard selection probabilities, for $5\times10^3$ rounds of communication, between which $40$ shard-local updates take place.
For both methods, we adopt fixed step-size $h = 10^{-4}$ and use mini-batches of size $m=50$.
The first $2 \times 10^4 $ samples were discarded for each method, and the remaining ones were thinned by two.

\textbf{Choice of $q_s$'s:} Similarly to the previous experiment, we computed the conducive terms by drawing three thousand samples independently from  densities proportional to the local likelihoods and imposing diagonal-multivariate normal approximations based on these samples.

\textbf{Results:} Table \ref{Tab:tab1} shows results in terms of average log-likelihood, evaluated on test and train data.
For the non-IID case, FSGLD significantly outperforms DSGLD on the test set. 
It is worth noting the low average log-likelihood achieved by DSGLD during training is a clear sign that it converged to models that do not generalize well.
For the IID case, both methods perform similarly.
In both cases, FSGLD shows a steep decrease in variance.
\begin{table}[h!]
\centering
\caption{\textbf{MLPs.} Average log-likelihood for MLPs learned with DSGLD and FSGLD. For the non-IID case, FSGLD clearly learns better models. For the IID case, both methods show similar performance.
Results reflect the outcome of ten repetitions (mean $\pm$ standard deviation). For all cases, FSGLD results in smaller standard deviation.}
\resizebox{0.475\textwidth}{!}{
\begin{tabular}{llccc}
\toprule
&   & Homogenous & Heterogenous &  \\
&   & (IID) & (non-IID) & \\
\midrule
\multirow{2}{*}{DSGLD}    & train & -0.69$\pm$0.001  & -0.44$\pm$0.062 &  \\
                          & test  & -0.70$\pm$0.001 & -1.11$\pm$0.312 &  \\ \hline
\multirow{2}{*}{FSGLD} & train & -0.69$\pm$0.00022 & -0.67$\pm$0.03 &  \\ 
                          & test   & -0.69$\pm$0.00022 & -0.67$\pm$0.03 & \\\bottomrule
\end{tabular}
}
\label{Tab:tab1}
\end{table}

\section{Related work}\label{sec:related}

\textbf{Variance reduction for stochastic optimization.}
Variance reduction has been previously explored to improve the convergence of stochastic gradient descent \citep[e.g.][]{Johnson013, defazio14, Defazio}. While federated optimization has gained increasing attention \citep{KonecnyMRR16}, variance reduction for distributed stochastic gradient descent has received only limited attention \citep{centralvr2016, daneaistats2020} so far. 

\textbf{Variance reduction for serial SG-MCMC.} 
Previous works by \cite{Dubey+others:2016,Baker2019} have proposed strategies to alleviate the effect of high variance in SGLD, for serial settings.
\cite{Dubey+others:2016} proposed two algorithms, SAGA-LD and SVRG-LD, both of which are based on using previously evaluated gradients to approximate gradients for data points not visited in a given iteration. The first algorithm, SAGA-LD, requires a record of individual gradients for each data point to be maintained. In the second one, SVRG-LD, the gradient on the entire data set needs to be periodically evaluated. The recently proposed SGLD-CV \cite{Baker2019} uses posterior mode estimates to build control variates, which are added to the gradient estimates to speed up convergence. 
Like SGLD-CV, our algorithm can be seen as a control variate method \citep{Ripley87}, but designed for distributed (federated) settings.

\textbf{Distributed/parallel SG-MCMC.} In the context of distributed SGLD, \cite{Li+others:2019_aaai} did some work proposing different communication protocols and analyzing their impact on convergence. Differently from our work, they do not focus on the impact of heterogeneity among data shards, which is key to federated settings.

\section{Conclusion}\label{sec:discussion}
In this work, we propose conducive gradients, a simple yet effective mechanism that leverages  approximations of each device's likelihood contribution and improves the convergence of SGLD for federated data.

We demonstrate i) that our method converges in scenarios where DSGLD fails, and ii) that it significantly  outperforms DSGLD in federated non-IID scenarios.
Additionally, our method can be seen as a variance reduction strategy for DSGLD. 
To the best of our knowledge, this is the first treatment of SG-MCMC for federated settings.

We also analyze the convergence of DSGLD to understand the influence of both data heterogeneity and device availability.
We also show convergence bounds for FSGLD and discuss how it supports our choice of local likelihood surrogates. 

Furthermore, given suitable surrogates $q_1, \ldots, q_S$, such as exponential family approximations, FSGLD can be simultaneously made efficient both in terms of memory and computation, imposing no significant overhead compared to DSGLD.
Nonetheless, we believe that, for very complex models, it could be useful to update the conducive terms as the chains navigate the posterior landscape.
We also leave open the  possibility of employing more expressive or computationally cheaper surrogates $\{q_s\}_{s=1}^{S}$, such as non-parametric methods or variational approximations. 

\begin{acknowledgements}
This work was funded by the Academy of Finland (Flagship programme: Finnish Center for Artificial Intelligence, FCAI, and grants 294238, 292334 and 319264). We are also grateful for the computational resources provided by the Aalto Science-IT Project.
\end{acknowledgements}

\bibliography{el-mekkaoui_652.bib}
\end{document}


\onecolumn
\maketitle
\appendix

\section{Background on convergence analysis for SGLD}\label{sec:background_supp}

\textbf{Regularity conditions.} Let $\psi$ be the functional that solves the Poisson equation $\mathcal{L}\psi =\phi - \hat{\phi}$.
Assume $\psi$ is bounded up to its third order derivative by a function $\Gamma$, such that $\|\mathcal{D}^{k} \psi\| \leq C_k \Gamma^{p_k}$ with $C_k, p_k > 0 \,\forall k \in \{0, \ldots, 3\}$ with $\mathcal{D}^{k}$ denoting the $k$th order derivative.
Assume as well that the expectation of $\Gamma$ w.r.t. $\theta_t$ is bounded ($\sup_t \mathbb{E} \Gamma^{p}[\theta_t] \leq \infty$) and that $\Gamma$ is smooth such that $\sup_{s \in (0,1)} \Gamma^{p}(s \theta + (1 - s) \theta^\prime) \leq C ( \Gamma^{p}(\theta) +   \Gamma^{p}(\theta^\prime))$, $\forall \theta, \theta^\prime$, $p \leq \max_k 2 p_k$, for some $C > 0$.

Under these regularity conditions, \cite{Chen:2015} showed the following result.

\begin{theorem}[See \cite{Chen:2015}] Let $U_t$  be an unbiased estimate of $U$, the unnormalized negative log posterior, and $h_t = h$ for all $t \in \{1, \ldots, T\}$.
Let $\Delta V_t = (\nabla U_t - \nabla U) \cdot \nabla$.
Under the assumptions above, for a smooth test function $\phi$, the MSE OF SGLD at time $K = hT$ is bounded for some $C>0$ independent of $(T, h)$ as:
    \begin{subequations}
    \begin{equation}
    \mathbb{E}[ (\overline{\phi} - \hat{\phi})^2 ] \leq C \left( \frac{\frac{1}{T} \sum_t \mathbb{E}[\| \Delta V_t \|^2]  }{T} + \frac{1}{T h} + h^2 \right) \label{eq:chen_bound}
    \end{equation}
    \end{subequations}
\end{theorem}

Equation (\ref{eq:chen_bound}) can also be written as:
\begin{subequations}
\begin{equation}
\mathbb{E}[ (\overline{\phi} - \hat{\phi})^2 ] \leq C \left( \frac{\frac{1}{T} \sum_t \mathbb{E}[\| \Delta V_t \psi(\theta_t) \|^2]  }{T} + \frac{1}{T h} + h^2 \right)
    \label{eq:is_this_chen}.
\end{equation}
\end{subequations}

For further analysis we add the assumption that $(\Delta V_t \psi(\theta))^2 \leq C^\prime \| \nabla U_t(\theta) - \nabla U(\theta) \|^2$ for some $C^\prime > 0$.
\section{Proof of Theorem 1: Convergence of DSGLD}\label{sec:th1}
Here, we follow the footprints of \citet{Chen:2015} later adopted by \citet{Dubey+others:2016}. Thus, we focus on bounding $\frac{1}{T} \sum_t \mathbb{E}[(\Delta V_t \psi(\theta_t) )^2]$, when $U_t(\theta_t) = v_{s_t}(\theta_t)$.  For some $C^\prime > 0$, we have:
\begin{subequations}
\allowdisplaybreaks
\begin{align}
\allowdisplaybreaks
     \frac{1}{C^\prime T} \sum_t \mathbb{E}[(\Delta V_t \psi(\theta_t) )^2]  &\leq \frac{1}{T} \sum_t \mathbb{E}[ \| \nabla U_t(\theta_t) - \nabla U(\theta_t) \|^2 ]\\
    &= \frac{1}{T}  \sum_t \mathbb{E} \left[ \left\| \frac{1}{f_s} \frac{N_s}{m} \nabla \log p(\x_{s_t}^{(m)} | \theta_t) - \nabla \log p(\x | \theta_t)  \right\|^2 \right]  \\ 
    %
    &= \frac{1}{T m^2} \sum_t \mathbb{E} \left[ \left\|  {\textstyle \sum}_{x_i \in \x_{s_t}^{(m)}} \frac{1}{f_s} N_s \nabla \log p(x_i | \theta_t) -  \nabla \log p(\x | \theta_t)  \right\|^2 \right] \label{step:before_tower_DSGLD}  \\ 
    &= \frac{1}{ m^2} \mathbb{E}_s  \mathbb{E}_{\x_{s_t}^{(m)} | s_t } \left[ {\textstyle \sum}_{x_i \in \x_{s_t}^{(m)}} \left\|   \frac{1}{f_s} N_s \nabla \log p(x_i | \theta_t) -  \nabla \log p(\x | \theta_t)  \right\|^2  \right] \label{step:tower_DSGLD}\\
    &\leq  \frac{1}{ m^2} \mathbb{E}_s\left[  \mathbb{E}_{\x_{s_t}^{(m)}| s_t} \left[  {\textstyle \sum}_{x_i \in \x_{s_t}^{(m)}} \Big\|   \frac{1}{f_s} N_s \nabla \log p(x_i | \theta_t) \Big\|^2 \right] \right] \label{step:bound_dsgld} \\
    &=  \frac{1}{ m^2} \mathbb{E}_s \left[ m \frac{N_s^2}{f_s^2}   \, \mathbb{E}_{x_i| s_t} \left[ \Big\|    \nabla \log p(x_i | \theta_t)  \Big\|^2 \right] \right]\\
    &\leq  \frac{1}{ m} \mathbb{E}_s \left[ \frac{N_s^2}{f_s^2}  \gamma_s^2 \right] = \frac{1}{ m} \sum_s f_s \frac{N_s^2}{f_s^2}  \gamma_s^2   = \frac{1}{m} \sum_s  \frac{N_s^2}{f_s}  \gamma_s^2 \label{step:final_dsgld}
    \end{align}
    \end{subequations}

Here, $\mathbb{E}_{\x_{s_t}^{(m)} | s_t}$ denotes that the expectation is taken w.r.t. a mini-batch of size $m$ with elements drawn with replacement and equal probability from shard $s_t$.
Expectations without explicit subscripts are taken w.r.t. all random variables.
To advance from Equation (\ref{step:before_tower_DSGLD}) to (\ref{step:tower_DSGLD}), we use law of iterated expectations and the fact that $\mathbb{E}[ \| \sum_i r_i \|^2 ] = \sum_i \mathbb{E}[ \|  r_i \|^2 ]$ for zero-mean independent $r_i$'s. 
To advance from Equation (\ref{step:tower_DSGLD}) to (\ref{step:bound_dsgld}), we use $\mathbb{E}[ \| r - \mathbb{E}[r]\|^2 ] \leq \mathbb{E}[ \| r \|^2 ]$.
Substituting Equation (\ref{step:final_dsgld}) in Equation (\ref{eq:is_this_chen}) yields the desired result. 
\section{Proof of Lemma 1:  Unbiasedness and finite variance}\label{sec:lm1}

Recall that the for the DSGLD update \citep{Ahn+others:2014} we have 
\[
    \mathbb{E}_{s, \x_{s_t}^{(m)}}\left[ \frac{1}{f_s} \frac{N_s}{m} \nabla \log p(\x_{s_t}^{(m)} | \theta_t)\right]  = \nabla \log p(\x | \theta_t).
\]
Furthermore, for \emph{conducive gradients} we have that:
\[
    \mathbb{E}_s \left[\nabla q(\theta_t) - \frac{1}{f_s} \nabla q_s (\theta_t) \right] = q(\theta_t) - \sum_s f_s \frac{1}{f_s} q_s(\theta_t) = 0.
\]
Since the FSGLD estimator is the sum of the DSGLD estimator and the \emph{conducive gradient}, it is unbiased. 

The sufficient condition for the DSGLD estimator to have finite variance is that the unnormalized log posterior is Lipschitz continuous. Similarly, since $q_1, \ldots, q_S$ are also Lipschitz continuous, their first derivatives are bounded, so the conducive gradient is a convex combination of bounded functions and has finite variance. Thus, their sum, the FSGLD estimator has finite variance.
\section{Proof of Theorem 2:  Convergence of FSGLD}\label{sec:th2}

We now bound $\frac{1}{T} \sum_t \mathbb{E}[(\Delta V_t \psi(\theta_t) )^2]$ for the FSGLD update equation, when $U_t(\theta_t) = v_{s_t}(\theta_t) + g_{s_t}(\theta_t)$.
\begin{subequations}
\allowdisplaybreaks
\begin{align}\allowdisplaybreaks
    \frac{1}{C^\prime T} \sum_t &\mathbb{E}[(\Delta V_t \psi(\theta_t) )^2] \nonumber\\
    &\leq \frac{1}{T} \sum_t \mathbb{E}[ \| \nabla U_t(\theta_t) - \nabla U(\theta_t) \|^2 ] \\
    &= \frac{1}{T m^2} \sum_t \mathbb{E} \left[ \Big\| \sum_{x_i \in \x_{s_t}^{(m)}} \frac{1}{f_s} \left( {N_s} \nabla \log p(x_i | \theta_t) -  \nabla \log q_s( \theta_t) \right)  + \nabla \log q(\theta_t) - \nabla \log p(\x | \theta_t)  \Big\|^2 \right]     \label{step:before_tower_cg} \\
    &=  \frac{1}{ m^2}  \mathbb{E}_s \mathbb{E}_{\x_{s_t}^{(m)} | s_t}  \left[ \sum_{x_i \in \x_{s_t}^{(m)}} \Big\| \frac{1}{f_s} \left( {N_s} \nabla \log p(x_i | \theta_t) -  \nabla \log q_s( \theta_t) \right)  +  \nabla \log q(\theta_t) - \nabla \log p(\x | \theta_t) \Big\|^2  \right]  \label{step:tower_cg}\\
    & \leq \frac{1}{ m^2}  \mathbb{E}_s \left[ \mathbb{E}_{\x_{s_t}^{(m)} | s_t} \left[  \sum_{x_i \in \x_{s_t}^{(m)}} \Big\| \frac{1}{f_s} \left( {N_s} \nabla \log p(x_i | \theta_t) -  \nabla \log q_s( \theta_t) \right)  \Big\|^2 \right] \right] \label{step:bound_cg}\\
     &= \frac{1}{m^2} \mathbb{E}_s\left[ m \frac{N_s^2}{f_s^2} \mathbb{E}_{x_i | s_t} \left[ \Big\|   \nabla \log p(x_i | \theta_t) -  N_s^{-1} \nabla \log q_s(\theta_t)   \Big\|^2  \right] \right]\\
     &= \frac{1}{ m^2} \sum_s f_s m \frac{N_s^2}{f_s^2}  \mathbb{E}_{x_i | s_t} \left[ \Big\|   \nabla \log p(x_i | \theta_t)  - N_s^{-1} \nabla \log q_s(\theta_t)   \Big\|^2  \right] \\
   &\leq \frac{1}{ m} \sum_s \frac{N_s^2}{f_s} \epsilon^2_s   \label{eq:th2_final}
   \end{align}
\end{subequations}
We proceed from Equation (\ref{step:before_tower_cg}) to (\ref{step:tower_cg}) using the law of iterated expectations and the fact that $\mathbb{E}[ \| \sum_i r_i \|^2 ] = \sum_i \mathbb{E}[ \|  r_i \|^2 ]$ for zero-mean independent $r_i$'s. To transition from Equation (\ref{step:tower_cg}) to (\ref{step:bound_cg}), we use $\mathbb{E}[ \| r - \mathbb{E}[r]\|^2 ] \leq \mathbb{E}[ \| r \|^2 ]$. The last line is obtained using \textit{Lemma} 2. We use this bound and Equation \ref{eq:is_this_chen} to get the desired result. 
\section{More details on Remark 3}

\begin{subequations}
Following \textit{Lemma} 2 and assuming that $\epsilon_s^2$ is a tight bound, i.e.
\begin{equation}
   \epsilon_s^2 \coloneqq \frac{1}{N_s}\sum_{x_i \in \x_s} \left\| \nabla \log p(x_i |\theta) - \frac{1}{N_s} \nabla \log q_s(\theta) \right\|^2 ,
\end{equation}
choosing $q_s$ that minimizes $\epsilon^2_s$ is equivalent to finding
\begin{equation}
    \min_{q_s} \max_\theta \frac{1}{N_s} \sum_{x_i \in \x_s} \left\| \nabla \log p(x_i |\theta) - \frac{1}{N_s} \nabla \log q_s(\theta) \right\|^2, \label{eq:def_eps}
\end{equation}
which is equal to
\begin{align}
\frac{1}{N_s} \min_{q_s} \max_\theta \sum_{x_i \in \x_s} \left[ \left\| \nabla \log p(x_i |\theta) \right\|^2 + \frac{1}{N_s^2} \left\|  \nabla \log q_s(\theta) \right\|^2 - \frac{2}{N_s} \left(\nabla \log q_s(\theta)\right)^\intercal \nabla \log p(x_i | \theta) \right],
\end{align}
and can be further developed into
\begin{align}
\frac{1}{N_s} \min_{q_s} \max_\theta  \left[ \sum_{x_i \in \x_s} \left\| \nabla \log p(x_i |\theta) \right\|^2 \right] + \frac{1}{N_s} \left\|  \nabla \log q_s(\theta) \right\|^2 -  \frac{2}{N_s} \left(\nabla \log q_s(\theta)\right)^\intercal \nabla \log p(\x_s | \theta).     
\end{align}

Completing the squares and using $\max a+b \leq \max a + \max b$, we get the following upper-bound for Equation \ref{eq:def_eps}:
\begin{equation}
    \frac{1}{N_s^2} \min_{q_s} \max_\theta \left[ \left\| \nabla \log q_s(\theta) - \nabla \log p(\x_s | \theta) \right\|^2  \right] + \frac{1}{N_s} \max_\theta \left[\frac{1}{N_s} \left\| \nabla \log p(\x_s | \theta) \right\|^2  + \sum_{x_i \in \x_s} \left\| \nabla \log p(x_i |\theta) \right\|^2 \right],
\end{equation}
in which only the first term depends on $q_s$.
\end{subequations}
\section{Additional experiments}\label{sec:moreexp}

In this section, we provide additional results for Bayesian linear regression. 
Since we can leverage the simple likelihood function and compute surrogates analytically, this setting is especially useful to understand the behavior of our method. 

\subsection{Linear regression}
In this set of experiments, we apply FSGLD to Bayesian linear regression and analyze its performance, which we measure in terms of MSE averaged over posterior samples.

\paragraph{Model}
The inputs of our model are $\boldsymbol{Z}=\left\{x_{i}, y_{i}\right\}_{i=1}^{N}$, where $x_i \in \mathbb{R}^d$ and $y_i \in \mathbb{R}$. 
The likelihood of the $i$th output $y_i \in\{0,1\}$, given the input vector $x_i$, is 
$p\left(y_{i} | x_{i}\right)=\mathcal{N}\left(y_i | \beta^{\top} x_{i}, \sigma_{e}\right)$, 
and we place the prior $p(\beta)=\mathcal{N}\left(\beta |\mathbf{0}, \lambda^{-1} I\right)$.

\paragraph{Setting}
We run experiments on three different datasets\footnote{Datasets can be downloaded from \url{https://archive.ics.uci.edu/ml/index.html}} from the UCI repository: Concrete ($1030$ samples, $9$ features); Noise ($1503$ samples, $6$ features); Conductivity ($17389$ samples, $81$ features)
We normalize and partition our datasets into (80\%) training and (20\%) test sets. 
In all our experiments, both DSGLD and FSGLD have the same hyper-parameters.
We sample $S=10$ disjoint data subsets for $r= 1000$ rounds each having  $600$ iteration per round, with fixed step-size $h_t = 10^{-5}$ and mini-batch size $m=10$.
All shards are chosen with same probability $f_1 = \cdots =  f_S = 1/S$.
We also burn-in the first ten thousand samples and thin the remaining  by a hundred. We set $q_s(\theta) = \mathcal{N}(\theta |\mu, \Sigma)$, with $\Sigma = (\x^{\top}\x)^{-1}$ and $\mu = (\sum y_i x_i) \Sigma^{-1}$, for each $s=1\ldots S$. 
We repeated the same experiment for $10$ different random seeds.
We report the average test MSE and its variance as a function of the number of posterior samples. 

\begin{figure}[tp]
    \centering
        \includegraphics[width=0.33\textwidth]{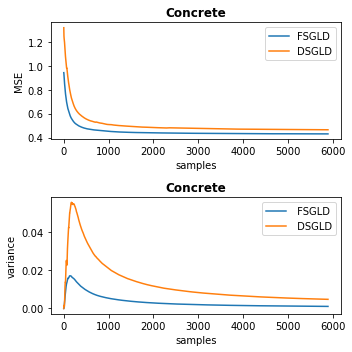}
        \includegraphics[width=0.33\textwidth]{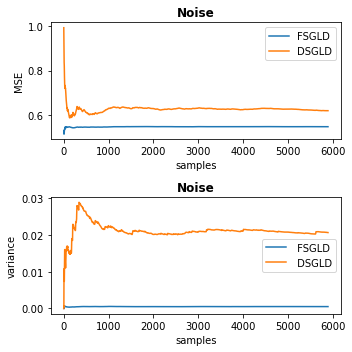}
        \includegraphics[width=0.33\textwidth]{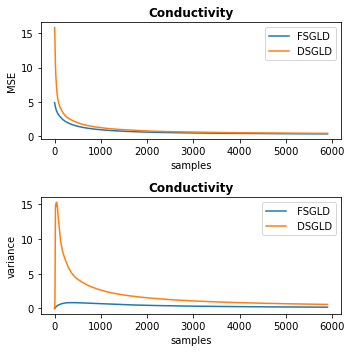}
    \caption{Average MSE and variance along time computed for DSGLD and FSGLD as a function of the number of samples. Overall, FSGLD converges to better performance than DSGLD. Additionally, FSGLD shows lower variance for all datasets.}
    \label{fig:LR_exp}
\end{figure}

\paragraph{Results}
Figure \ref{fig:LR_exp} shows the cumulative MSE and its variance. 
Overall, FSGLD converges faster than DSGLD in MSE, with the notable exception of the Conductivity dataset, for which both methods converge virtually at the same time.
In the case of the Noise dataset, FSGLD additionally converges to a much lower MSE.
Notably, our method also presents clearly lower variance for all datasets.

\subsection{Metric learning} \label{appendix:laplace}

\begin{figure}[H]
    \centering
    \includegraphics[width=0.33\textwidth]{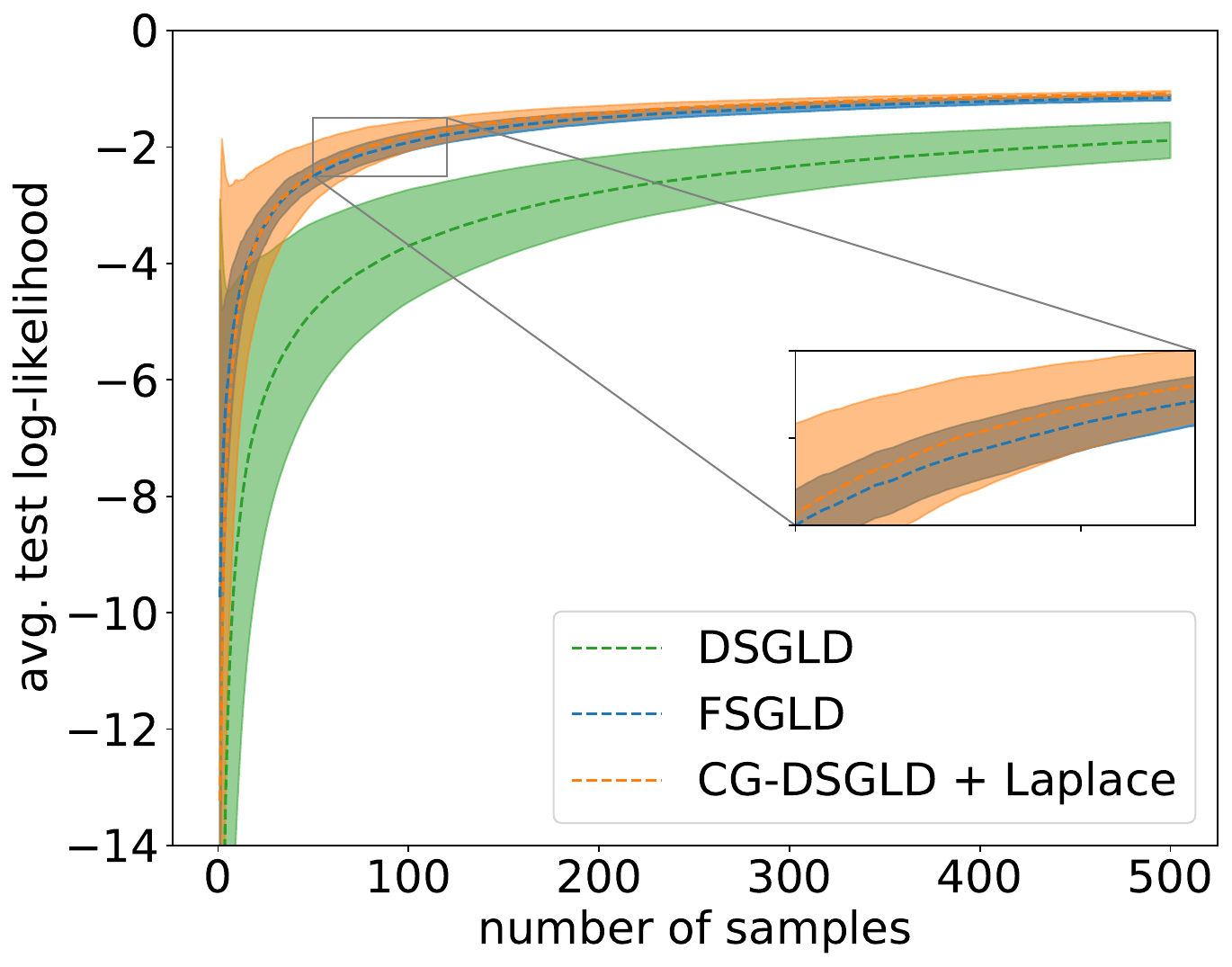}
    \caption{ In avgerage, CG-DSGLD with Laplace or MCMC-based $q_s$'s perform on par. Notably, MCMC yields smaller variance in the metric learning experiment.}
    \label{fig:my_label}
\end{figure}

While we employed MCMC-based surrogates for $q_1,\dots,q_S$, we can also use FSGLD with coarser approximations.
As an example, we also run FSGLD on the metric learning posterior of \autoref{exp:metric} using Laplace approximations.
Notably, \autoref{fig:my_label} shows that both options lead to similar average results, but Laplace approximations result in higher variance. 

\bibliographystyle{natbib}
\bibliography{uai2021-template.bib}